\documentclass[journal]{IEEEtran}

\usepackage[latin1]{inputenc}
\usepackage[T1]{fontenc}
\usepackage{cite}
\usepackage[pdftex]{graphicx}
\usepackage[cmex10]{amsmath}
\usepackage{dblfloatfix}
\usepackage[tight,footnotesize]{subfigure}
\usepackage{url}
\usepackage[pdftex]{hyperref}

\begin{document}

\title{Spatial Neural Networks and their Functional Samples: Similarities and Differences}

\author{Lucas~Antiqueira and Liang~Zhao
\thanks{This work was supported by CNPq (Brazil) under grants 383720/2011-7 and 150550/2012-0.}
\thanks{L. Antiqueira is with the Institute of Mathematical and Computer Sciences, University of São Paulo, São Carlos, SP, Brazil, e-mail: lantiq@icmc.usp.br.}
\thanks{L. Zhao is with the Faculty of Philosophy, Sciences and Languages of Ribeirão Preto, University of São Paulo, Ribeirão Preto, SP, Brazil, e-mail: zhao@usp.br.}
}

\maketitle

\begin{abstract}
Models of neural networks have proven their utility in the development of learning algorithms in computer science and in the theoretical study of brain dynamics in computational neuroscience. We propose in this paper a spatial neural network model to analyze the important class of \emph{functional networks}, which are commonly employed in computational studies of clinical brain imaging time series. We developed a simulation framework inspired by multichannel brain surface recordings (more specifically, EEG -- electroencephalogram) in order to link the mesoscopic network dynamics (represented by sampled functional networks) and the microscopic network structure (represented by an integrate-and-fire neural network located in a 3D space -- hence the term \emph{spatial neural network}). Functional networks are obtained by computing pairwise correlations between time-series of mesoscopic electric potential dynamics, which allows the construction of a graph where each node represents one time-series. The spatial neural network model is central in this study in the sense that it allowed us to characterize sampled functional networks in terms of what features they are able to reproduce from the underlying spatial network. Our modeling approach shows that, in specific conditions of sample size and edge density, it is possible to precisely estimate several network measurements of spatial networks by just observing functional samples.
\end{abstract}

\begin{IEEEkeywords}
Computer simulation, Modeling, Network theory (graphs), Spatial networks.
\end{IEEEkeywords}

\section{Introduction}

\IEEEPARstart{M}{any} developments in the so-called \emph{network science}~\cite{Brandes2013} have been successfully applied to empirical studies of natural, sociological and technological systems~\cite{Newman2010}. Nevertheless, network analyses are frequently based on sampled data, which imposes an inherent problem of data incompleteness (missing or erroneous nodes/edges). The sampling problem, although fundamental, has been insufficiently understood in the case of network data (some notable efforts are~\cite{Achlioptas2009,Dall'Asta2006a,Latapy2008,Stumpf2005}). In this paper we report a comprehensive simulation framework for the study of functional networks, an important class of network samples, usually taken from brain imaging methods~\cite{Stam2010, Rubinov2011, Palva2012}. We have used a spatial (neurons and synapses are distributed in a 3D space) neural network to investigate the implications of the functional sampling process, taking as inspiration experimental studies frequently done in neuroinformatics that heavily depend on multichannel recordings (in our case, the entire model is based on the electroencephalogram - EEG). This type of functional network is experimentally constructed by means of pairwise cross-correlations (or other time series measurement) computed between all EEG signals~\cite{Stam2010, Rubinov2011, Palva2012}. The fundamental problem being addressed here is that of relating the microscopic organization patterns (spatial network) with the mesoscopic dynamical patterns (functional network) using well known network/graph measurements.

Some critical analyses of functional brain sampling have already been performed. It has been shown, for instance, that functional networks obtained from single-neuron spike trains recorded in the monkey visual system tend to overestimate the small-world effect~\cite{Gerhard2011} -- a small-world network is locally clustered while distances between nodes are short on average~\cite{Watts1998}. In other study, the number of recorded signals have been evaluated in the light of network stability~\cite{Joudaki2012}. More specifically, the authors used experimental human EEG data and analyzed the dependence of network measurements on the sample size (number of electrodes). Results showed that, among other findings, larger functional samples show an increase in efficiency and assortativity. On the other side, functional networks obtained from human long-term intracranial EEG signals revealed a recurring functional network core that persisted independently of cognitive process~\cite{Kramer2011}. It has also been shown that EEG-like functional samples tend to incorporate random graph properties when the underlying neural networks resemble a small-world~\cite{Antiqueira2010a}. Although conclusions are relevant, those simulations are based on techniques that overlook neural dynamics. We instead propose to use a more realistic simulation approach based on a spiking neuron model. Moreover, we also point out other theoretical study reported in~\cite{Shkarayev2009}, which was based on the integrate-and-fire model and mean-field theory. Authors came to an interesting conclusion: it is possible to generate scale-free functional samples from scale-free neural networks -- in a few words, a scale-free network has a power-law degree distribution~\cite{Barabasi1999}. We report in the following sections results based on a wide range of different network measurements that expands conclusions beyond the analysis of degree distributions of scale-free networks -- without even assuming that biological neural networks are purely scale-free.

\begin{figure*}[!t]
	\centering
	\includegraphics[width=0.65\linewidth]{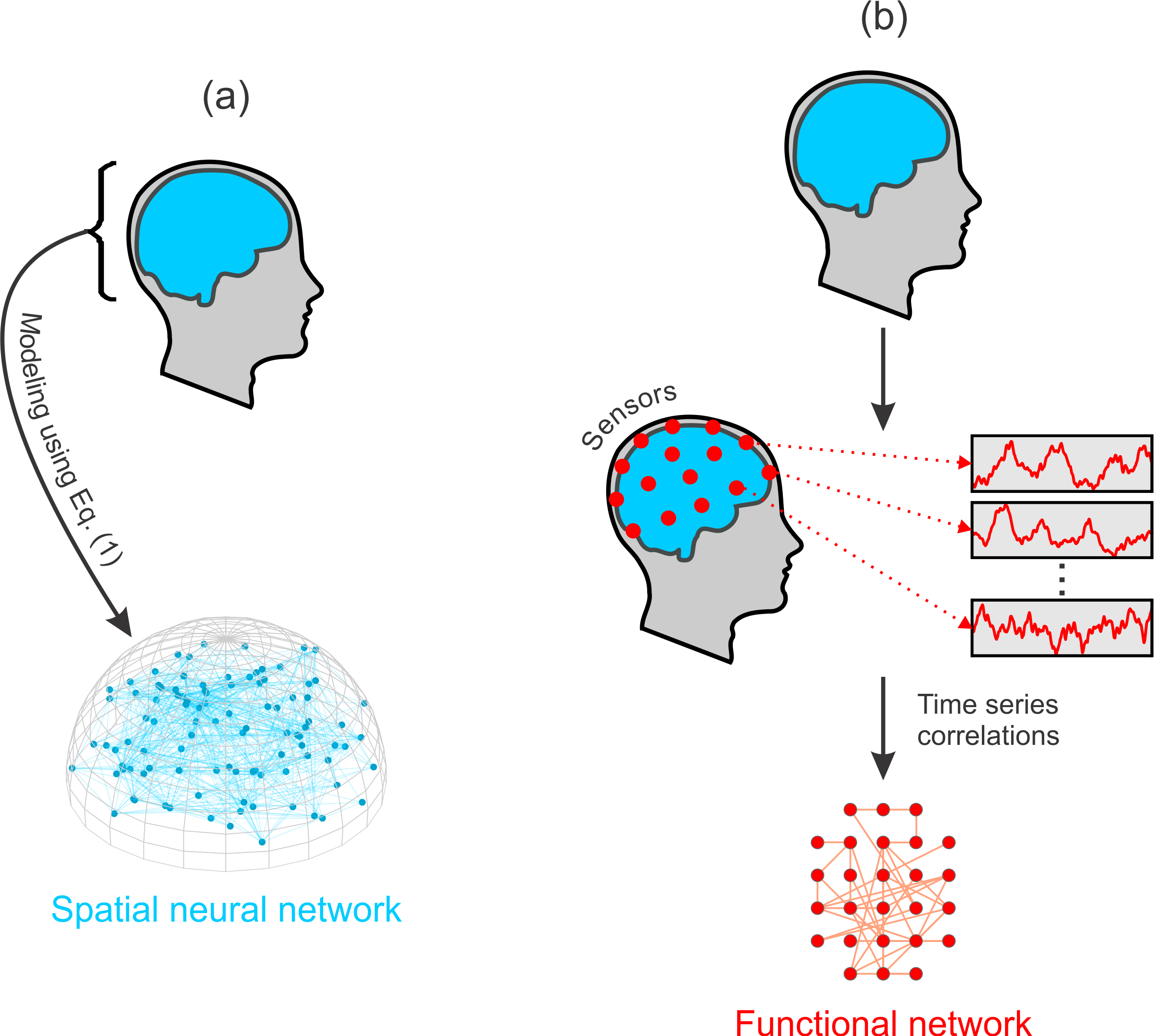}
	\caption{Illustration of the general methodology adopted in this paper. (a)~Structure of neurons and synapses in the human brain are modeled as a \emph{spatial network} where nodes are placed inside a semi-sphere in order to represent finite size effects (details in Section~\ref{sec:neuralnet}). (b)~Experimental method used in clinical studies of functional networks, where sensors (electrodes in the case of EEG) are placed over the human scalp in order to obtain one time series of electric potentials for each sensor. Cross-correlations between all time series are computed, where the strongest correlations lead to the creation of edges in the so-called \emph{functional network} (each functional node represents one time series/sensor). Integrate-and-fire neurons were incorporated in the spatial network of (a) in order to build \emph{pseudo-EEG} signals (details in Section~\ref{sec:func}) and therefore mimic the procedure shown in (b).}
	\label{fig:schematic}
\end{figure*}

The simulation technique reported in this paper was defined in order to mimic experimental situations, and mainly encompass (i)~a nonuniform network model that incorporates neuroanatomical connectivity properties (spatial network, Fig.~\ref{fig:schematic}a), (ii)~the construction of mesoscale cortical electrical signals (\emph{pseudo-EEG}) based on the integrate-and-fire model, and (iii)~the estimation of functional networks based on those signals. Steps (ii) and (iii) correspond the experimental procedure depicted in Fig.~\ref{fig:schematic}b. A large set of network (graph) measurements is computed for each spatial and functional network in order to perform detailed comparisons between them. Examples of measurements are those based on shortest-paths (betweenness, average distance, efficiency, ect) and those relying on local connectivity (clustering coefficient, assortativity, etc). We also computed concentric (or hierarchical) generalizations of local measurements~\cite{Costa2005a}. Results for a range of sample sizes and densities indicate that some connectivity properties of spatial networks are well reflected in functional samples (e.g., closeness vitality). Other network features heavily depend on the number of nodes and/or edge density (e.g., node and edge betweenness). These findings pave the way for other similar theoretical developments using spiking neuron models and spatial networks.

This paper is organized as follows. Next (Section~\ref{sec:methods}) we describe and justify the simulation methods adopted, and also define the measurements employed for network comparison. A detailed analysis of estimated functional samples is carried out in Section~\ref{sec:results}. Comments on further directions for this research as well as other concluding remarks are given in Section~\ref{sec:concl}.

\section{Material and Methods}
\label{sec:methods}

The following subsections contain details on the organization of the spatial network (Section~\ref{sec:neuralnet}) and the simulation procedure adopted to obtain functional networks (Section~\ref{sec:func}). The latter encompass details about the spiking neuron model (integrate-and-fire), the simulated signals inspired by EEG experiments, and the method used to build functional networks from those signals. All the measurements employed to compare networks are defined accordingly (Section~\ref{sec:measur}). Finally, we provide details about the specific parameters and organization of the simulation experiment (Section~\ref{sec:setup}).

\subsection{Generating Spatial Networks}
\label{sec:neuralnet}

The establishment of synapses in the spatial network is governed by a probabilistic rule that yields higher probabilities of connection between spatially closer neurons than between distant neurons. In other words, global connections (shortcuts) are created less frequently than local connections, therefore incorporating the concept of wiring economy -- long-range synapses impose higher costs in terms of volume and transmission~\cite{Bullmore2012}. The probability of creating a synapse from neuron $i$ to neuron $j$ is given by the following equation:
\begin{equation}
	p(i,j) = \beta \exp \left[ - \alpha d(i,j) \right]~,
	\label{eq:prob}
\end{equation}
where $\beta$ and $\alpha$ are parameters that shape the interplay between local and global connections ($0 < \beta \leq 1$, $\alpha > 0$). The Euclidean distance between neurons $i$ and $j$ is represented by $d(i,j)$. We consider for simplicity that $N$ neurons are randomly distributed inside a volume (a unit radius semi-sphere) that represents the finite size effects of the human skull (see an example in Fig.~\ref{fig:schematic}a). Equation~\eqref{eq:prob} is derived from developments previously made on \emph{spatial networks}~\cite{Waxman1988,Kaiser2004a}.

It is worth to mention that this model generates \emph{directed} networks, i.e., each synapse $(i,j)$ connects a presynaptic neuron~$i$ and a postsynaptic neuron~$j$, which is in fact a neuroanatomical property. Consequently, although we have $p(i,j) = p(j,i)$ in \eqref{eq:prob}, the creation of synapse $(i,j)$ in this model does not depend on the existence of synapse $(j,i)$ (and vice-versa).

\subsection{Generating Functional Networks}
\label{sec:func}

\subsubsection{Integrate-and-Fire Neuron Model}

The spatial network model of the previous section is static in the sense that it does not incorporate any neuronal dynamics, which is in fact necessary to conceptually connect the structural level of neurons/synapses to the mesoscopic functional level. More specifically, in order to construct EEG-like signals (more details in Section~\ref{sec:eeg}) we have to incorporate the time varying membrane potential of each neuron. Spiking neuron models emerge as a natural choice, since spiking dynamics is a essential feature of biological neurons that directly shapes postsynaptic membrane potentials. Many types of models could be employed, ranging from simple neurons to the complex and more realistic ones~\cite{Trappenberg2002}. We chose the \emph{leaky integrate-and-fire} (LIF) model which, despite showing a simple behavior when considered in isolation, enables complex dynamics inside nonuniform networks and features enough anatomically and physiologically inspired features for our simulations. Moreover, we focus here on the system dynamics rather than on the specific shape and function of specific individual neurons. 

This neuron model is represented by an RC (resistor-capacitor) parallel circuit that shapes neuron dynamics, which comprises the time variation of the membrane potential and spike (action potential) trains~\cite{Trappenberg2002, Deco2008}. A differential equation defines the variation of the membrane potential $V_m$ of one neuron:
\begin{equation}
	\tau_m\frac{dV_m(t)}{dt} = - [V_m(t) - V_{rest}] + RI(t)~.
	\label{eq:diff}
\end{equation}
The membrane potential $V_m$ is the potential difference across the capacitor (expressed in volts -- V), which represents the difference between the neuron internal potential and the potential of the external medium (hence negative values~\cite{Trappenberg2002}). Other parameters are: the resistance of the circuit $R$ in ohms ($\Omega$), the input current $I$ in amperes (A), the neuron resting potential $V_{rest}$ (represented by a battery in the model circuit), and the membrane time constant (in seconds) $\tau_m = RC$ ($C$ is the capacitance in farads -- F).

The membrane potential is entirely driven by \eqref{eq:diff} until a threshold $\theta$ is reached, which causes the model to generate a pulse (neuron spike). The two following equations specify the instant of a spike $t^{(s)}$ and the reset procedure of the membrane potential to a predefined value ($V_{reset}$) after a spike, respectively:
\begin{equation}
	V_m(t^{(s)}) = \theta~,
\end{equation}
\begin{equation}
	\lim_{\delta \rightarrow 0} V_m(t^{(s)} + \delta) = V_{reset}~.
\end{equation}
The model also specifies a time varying input current $I$ that depends on spikes generated by presynaptic neurons (now taking into account the spatial network structure of Section~\ref{sec:neuralnet}):
\begin{equation}
	I(t) = \sum_j \sum_{t^{(s)}_j} w_j\alpha\left(t-t^{(s)}_j\right)~,
	\label{eq:curr}
\end{equation}
where $j$ iterates over all presynaptic neurons and $t^{(s)}_j$ iterates over all spikes generated by $j$. The strength or efficiency of the synapse is denoted by $w_j$ ($w_j > 0$ for excitatory and $w_j < 0$ for inhibitory synapses~\cite{Trappenberg2002}). In \eqref{eq:curr} currents are shaped by an $\alpha$-function as follows~\cite{Roth2009}: 
$$ \alpha(x) = \frac{x}{\mu}\exp\left(1-\frac{x}{\mu}\right)~, $$ 
with maximum value $1$ at $x = \mu$ (the growth time is therefore specified by $\mu$). Variable $x$ is substituted in \eqref{eq:curr} by the elapsed time since a spike occurred in a presynaptic neuron $j$ (each spike is considered separately).

The LIF model is therefore completely defined by~\eqref{eq:diff}--\eqref{eq:curr}, plus the $\alpha$-function. Moreover, given that biological neurons cannot fire immediately after a previous spike~\cite{Trappenberg2002}, it is included in the model a refractory period $t_{ref}$ during which $V_m$ stays equal to $V_{reset}$ after a spike.

\subsubsection{Pseudo-EEG Signals}
\label{sec:eeg}

\begin{figure}[!t]
	\centering
	\includegraphics[width=0.95\columnwidth]{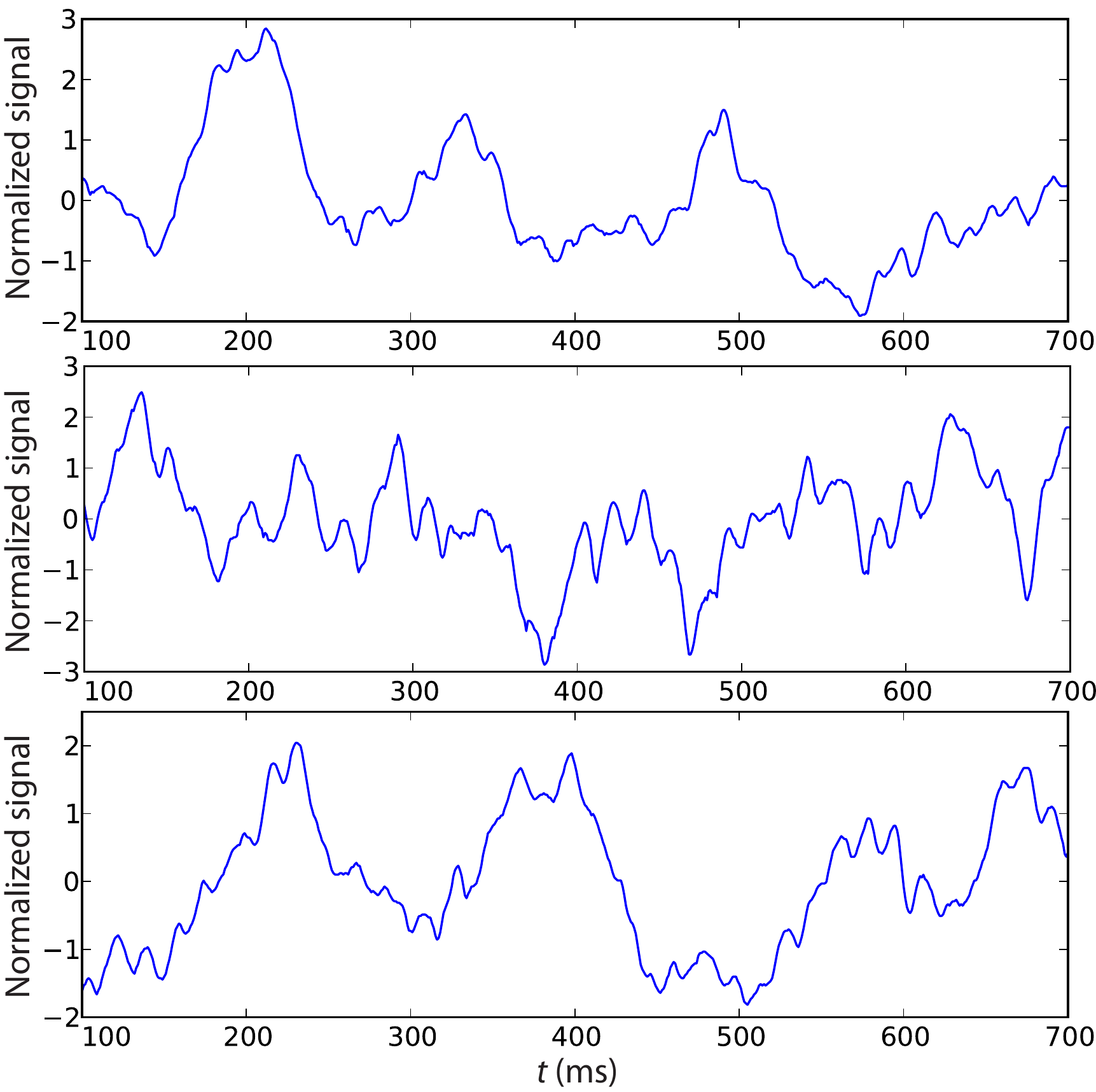}
	\caption{Three examples of normalized (z-score) surface recordings (\emph{pseudo-EEG}) for a single network simulation. Different surface positions were chosen for each of the simulated signals. Simulation details are given in Section~\ref{sec:setup}.}
	\label{fig:eeg}
\end{figure}

In order to simulate signals inspired by conventional electroencephalography we consider that a set of $N_f$ recording points (functional sampling points or just ``sensors'') is uniformly distributed over the volume that contains the spatial network of $N$ nodes. By uniform we mean that all sensors are equidistant from each other. Then the mesoscale electric potential $S_{i_f}(t)$ recorded by sensor $i_f$ at time $t$ is an attenuated sum of individual membrane potentials of neurons:
\begin{equation}
	S_{i_f}(t) = \frac{1}{N}\sum_{i=1}^{N} \frac{V_m(i,t)}{d(i_f,i)^2}~,
	\label{eq:eeg}
\end{equation}
where the membrane potential of neuron $i$ in time $t$ is denoted by $V_m(i,t)$. Recall that sensor $i_f$ is located at the surface of the semi-sphere, whereas neuron $i$ is placed inside the semi-sphere, therefore $d(i_f,i)$ denotes the Euclidean distance between them. For the purpose of simulating (approximately) real distances, we consider in \eqref{eq:eeg} that the semi-sphere has a radius of 200~mm (units are in this case given in mm since membrane potentials are in the order of mV). In our simulations the medium between neurons and electrodes has constant conductivity. Fig.~\ref{fig:eeg} shows three examples of mesoscale simulated potentials.

We use an inverse square rule to attenuate membrane potentials. Therefore, only neurons close to the sensor significantly contribute to the formation of the captured signal. The reasoning behind this behavior is that neuron potentials can be approximated by dipoles (two poles with opposite charges) in the case of scalp encephalographic measurements, which allows the use of a inverse quadratic decay~\cite{Nunez2006}. Moreover, in order to complement our reasoning behind \eqref{eq:eeg}, we note that the signal captured by real EEG recordings is mostly influenced by microscopic postsynaptic currents~\cite{Olejniczak2006, Buzsaki2012} (hence postsynaptic potentials).

\subsubsection{Building Functional Edges}

One of the tools frequently applied to the analysis of neural multichannel experiments is the cross-correlation between all recorded time series. Take any pair of z-score normalized pseudo-EEG signals $S_{i_f}(t)$ and $S_{j_f}(t)$ (obtained from surface sensors $i_f$ and $j_f$, respectively), then the cross-correlation between them is computed as follows~\cite{Pereda2005}:
\begin{equation}
	C_{i_f,j_f}(\tau) = \frac{1}{N_s - \tau} \sum_{k=1}^{N_s - \tau} {S_{i_f}(k+\tau)S_{j_f}(k)}~.
	\label{eq:correl}
\end{equation}
The number of recorded points in both signals is denoted by $N_s$, whereas the time lag of correlation is $\tau$. This formula gives results between $-1$ and $1$, but we follow the usual approach of taking the maximum absolute value for a given time lag interval.

A new network is then constructed as follows. Let $W$ be the symmetric correlation matrix between all $N_f$ recorded signals. This square matrix naturally represents an undirected weighted graph with all possible edges, where nodes represent sensors. The \emph{functional network} is constructed by thresholding $W$, i.e., an adjacency matrix $A$ is created with $A(i,j)=1$ if and only if $W(i,j)$ is above a given threshold ($A(i,j)=0$ otherwise). As in graph theory, the adjacency matrix $A$ completely describes an unweighted and undirected functional network (graph). Consequently, only the strongest correlations lead to the creation of edges in the functional sampled network. Thresholds are usually arbitrarily chosen, but we specified one that allows a matching between a relevant property of both the underlying spatial network and the functional network. More specifically, we always choose a threshold that yields a functional network with the same edge density as the spatial network. Density is the proportion of edges in a network (the number of actual edges divided by the total possible number of edges).

\subsection{Network Measurements}
\label{sec:measur}

We are now able to compare functional networks and their respective spatial networks. To perform this comparison we resource to a rich set of network measurements, which have been defined for the purpose of characterizing specific network features. Since we compare undirected (functional) to directed (spatial) networks, we have to consider that each undirected functional edge is in fact a pair of two symmetric directed edges. Note that this is just a conceptual adaptation where no network information is lost. Measurement definitions that follow are thus based on directed edges encoded, whenever necessary, in an adjacency matrix $A$.

\subsubsection{Maximum degree}
The degree can be thought as the simplest network measurement, which is just the number of edges attached to a given node $i$. For directed networks there is a distinction between in- and out-degrees (in- and out-going edges, respectively): $k_{in}(i) = \sum_{j=1}^{N}A(j,i)$ and $k_{out}(i) = \sum_{j=1}^{N}A(i,j)$. We are interested here in analyzing measures of the whole network instead of single nodes. For that purpose the average network degree is usually taken. Since the average degree will always be the same in our comparisons (see Section~\ref{sec:setup}, where we explain that comparisons are performed between networks of same size and density), we have chosen to use the maximum in-degree (and out-degree) value of a given network.

\subsubsection{Clustering coefficient}
This coefficient has been proposed to study small-world networks with undirected edges~\cite{Watts1998}, and it quantifies the proportion of local clustering of a network. Here we devised a way to compute this coefficient in directed networks. Let $n(i)$ be the number of neighbors of node $i$ (it does not matter if they are in- or out-going neighbors) and $e(i)$ be the number of directed edges that connect those neighbors among themselves (i.e., edges connecting to $i$ are not taken into account). The clustering coefficient $C(i)$ of node $i$ is equal to:
\begin{equation}
	C(i) = \frac{e(i)}{n(i) \left( n(i)-1 \right) }~.
\label{eq:ccoeff}
\end{equation}
This measurement quantifies the level of interactions among neighbors of $i$: if neighbors are completely interconnected, then $C(i)=1$; on the opposite side, if those neighbors have no link between them, then $C(i)=0$. We consider the average clustering coefficient of a network as a measure of intensity of its local node grouping.

\subsubsection{Assortativity}
One way to quantify degree correlations is by calculating the assortativity index~\cite{Newman2002}. This measurement is useful to check if, for instance, highly connected nodes tend to create links among each other. It can be calculated using the Pearson correlation coefficient on the degrees of both ends of every edge. More specifically, for every directed edge $(i,j)$ we check if the in-degree of $i$ is linearly correlated with the in-degree of $j$. We call this measurement the ``in-in'' assortativity coefficient. We also consider the other three possible combinations of in- and out-degrees at both ends of every edge: out-out, in-out, and out-in, which consequently leads to three more assortativity indexes. The interpretation of this coefficient is naturally the same as for the Pearson correlation coefficient, where absolute values close to 1 mean strong linear correlation, whereas values close to 0 represent weak or no correlations. Positive or negative correlations are indicated by the sign of the coefficient.

\subsubsection{Shortest path length}
The distance $d(i,j)$ (or length of the shortest path) between nodes $i$ and $j$ is frequently used to evaluate the overall node separability. We take the distances between all possible pairs of nodes to compute the average shortest path length of a network: 
\begin{equation}
	d = \frac{1}{N(N-1)} \sum_{i=1}^{N}\sum_{j=1}^{N} d(i,j)~.
\end{equation} 
If $i$ and $j$ are placed inside different connected components we take $d(i,j) = N$, since conceptually it is an infinite distance. Note that for directed graphs $d(i,j)$ is not necessarily equal to $d(j,i)$, thus strictly speaking it is not a distance measure. Nevertheless, we use the term distance for simplicity.

\subsubsection{Global efficiency}
To avoid artifacts in the case of unconnected nodes we have also considered the measurement of network efficiency~\cite{Latora2001}, which is the average of the inverse of distance:
\begin{equation}
	E = \frac{1}{N(N-1)} \sum_{i=1}^{N}\sum_{\underset{j\neq{}i}{j=1}}^{N} \frac{1}{d(i,j)}~.
\end{equation}
The inverse of an infinite distance is therefore taken as 0. When distances tend to be short efficiency approaches 1 (a complete graph has maximum efficiency, for instance), whereas for higher distances smaller efficiencies appear (the extreme case of zero efficiency is a totally unconnected graph, i.e., with no edges).

\subsubsection{Betweenness centrality}
Another measurement based on shortest paths is betweenness~\cite{Freeman1977}. It estimates the probability of finding a node (or edge) inside a shortest path. More specifically, the betweenness centrality of a node $i$ is given by:
\begin{equation}
	B(i) = \sum_{a=1}^{N}\sum_{\underset{b\neq{}a}{b=1}}^{N} \frac{\sigma(a,i,b)}{\sigma(a,b)}~,
	\label{eq:bet}
\end{equation}
where $\sigma(a,b)$ is the number of different paths of shortest length that connect nodes $a$ and $b$, and $\sigma(a,i,b)$ is the number of those paths that include node $i$. We take the average of all $B(i)$ as a measurement of betweenness for an entire network. Finally, if we consider that $i$ is an edge instead of a node, the interpretation of \eqref{eq:bet} is naturally adapted and thus we also have a measurement of edge betweenness centrality.

\subsubsection{Closeness vitality}
This measurement has been proposed to quantify the effects of excluding a node from a network~\cite{Koschutzki2005}. To define it, let $\mu$ be the sum of all shortest path lengths ($\sum_{i=1}^{N}\sum_{j=1}^{N} d(i,j)$). Then consider that we remove from the network a node $i$ along with its corresponding edges and recompute the sum of path lengths (let us denote this new summation by $\mu_i$). The closeness vitality of $i$ is:
\begin{equation}
	V(i) = \mu_i - \mu~,
\end{equation}
which is the total increase in all shortest path lengths when $i$ is discarded. As an overall closeness vitality measurement (i.e., for an entire network) we take the average of all individual $V(i)$.

\subsubsection{Concentric number of nodes}
The class of concentric (or hierarchical) measurements is based on the notion of higher order neighborhoods of a given node $i$, not just its immediate neighbors~\cite{Costa2004b}. Let us consider that the level $h=0$ of concentric neighborhoods is composed of just node $i$. The next level ($h=1$) if formed by those nodes directly connected to~$i$, independently of edge direction (i.e., in-going or out-going edges are treated in the same way). Next, level $h=2$ contains nodes directly connected to nodes of level $h=1$ that are not already included in the first level. Generally, level $h > 0$ is composed of nodes not yet associated with any level that are directly connected to any node of level $h-1$. With this definition, levels must be computed for increasing $h$, not in any order. Breadth-first search is sufficient to find those levels given that (for our purposes) edge directions are removed from directed networks. Finally, the first concentric measurement applied to our work is the concentric number of nodes of a node $i$ at level $h$, which is just the number of nodes associated with level $h$ given that the ``center'' node is $i$. We compute this measurement for levels 2, 3, and 4. For every concentric measurement (including the next) we take the average between all nodes as a global network index of hierarchical connectivity.

\subsubsection{Concentric degree}
Connections between different concentric levels can be used to generalize in- and out-degrees. In a short definition, the in-degree $k_{in,h}(i)$ of node $i$ at level $h$ is the number of edges coming from level $h$ to level $h-1$. Notice that the in-degree $k_{in,1}(i)$ at level $1$ is equivalent to the usual in-degree $k_{in}(i)$. The concentric out-degree $k_{out,h}(i)$ is defined analogously: it is the number of edges coming from level $h-1$ to level $h$ (again, $k_{out,1}(i) = k_{out}(i)$). In our analyses we consider concentric in- and out-degrees at levels 2 to 4.

\subsubsection{Concentric neighbor degree}
Another concentric measure is the neighbor average degree, which quantifies the overall connectivity of nodes at level $h$. The concentric neighbor in-degree $kn_{in,h}(i)$ of node $i$ at level $h$ is defined as follows, where $L_{h}(i)$ is the set of nodes (with $n_{h}(i)$ elements) at level $h$ of node $i$:
\begin{equation}
	kn_{in,h}(i) = \frac{1}{n_{h}(i)} \sum_{j \in L_{h}(i)} k_{in}(j)~.
\end{equation}
The concentric neighbor average out-degree $kn_{out,h}(i)$ is defined analogously. We compute these measurements for levels 1 to 4.

\subsubsection{Concentric clustering coefficient}
An interesting concentric generalization can be done in the case of the clustering coefficient. Let $e_h(i)$ and $n_h(i)$ be the number of directed edges and nodes, respectively, at the level $h$ of $i$. For computing the number of edges $e_h(i)$ take into consideration only edges connecting nodes inside level $h$. The concentric clustering coefficient $C_h(i)$ of node $i$ at level $h$ is equal to:
\begin{equation}
	C_h(i) = \frac{e_h(i)}{n_h(i) \left( n_h(i)-1 \right) }~.
	\label{eq:hccoeff}
\end{equation}
The usual clustering coefficient $C(i)$ is therefore equal to $C_1(i)$. Levels 2 to 4 were considered in our simulations.

\subsection{Simulation Setup}
\label{sec:setup}

Simulations were implemented in Python using the following packages: NetworkX~\cite{Hagberg2008}, Brian~\cite{Goodman2008}, and PyNN~\cite{Davison2009}. Spatial networks of $N=2000$ nodes were created with $\alpha = 2$. The other connection probability parameter ($\beta$) was set to $0.3$, $0.4$, and $0.5$, each time generating other 20 spatial networks. Therefore we have three sets of spatial networks with distinct (average) edge densities induced by those different $\beta$: lower (6.5\%), intermediate (8.6\%), and higher densities (10.8\%). In this manner we are able to assess the influence of network density on functional sampling.

Integrate-and-fire neurons have their resting, reset and threshold potentials set to $V_{rest} = -70$~mV, $V_{reset} = -75$~mV, and $\theta = -55$~mV, respectively. Their membrane capacitance $C$ is equal to $0.5$~nF, whereas the growth time $\mu$ of the $\alpha$-function is 5~ms. Their membrane time constant $\tau_m$ and refractory period $t_{ref}$ were both set to 15~ms. Some values ($V_{rest}$ and $\theta$) were inspired by actual neurophysiological observations~\cite{Martini2011}, although for the other variables we needed to search the parameter space for specific values that prevent sub or super neural activity (absence of spikes or overexcitability, respectively).

Parameter variability was reserved for synaptic weights: excitatory weights follow a Gaussian distribution with average~$1.0$ and standard deviation $0.1$ ($-5.0$ and $0.5$, respectively, for inhibitory weights, which comprise 20\% of all synapses). Total simulation time is $3000$~ms with time steps of 1~ms. The system input is a set of Poisson spike generators individually connected to a fraction of neurons (2\% of $N$). Each generator has an expected number of 20 spikes per second. Pseudo-EEGs were calculated after the initial 100~ms of simulation time (i.e., we removed the transient period of membrane potential dynamics). Cross-correlations were computed for time lags $\tau$ between -50 and 50~ms.

Seven different number of sensors $N_f$ were considered (40, 50, \ldots, 100) for multichannel surface recording. Given one specific $N_f$, one functional network is obtained for each spatial network. To properly compare two networks (in our case, functional and spatial) by means of network measurements, it is necessary both have the same number of nodes and edges. Nevertheless, since EEG has a low spatial resolution, the size of a functional network is much smaller than the size of the underlying spatial network. We are now able to restate the main purpose of this work in the form of a question: Can simulated functional networks resemble the connectivity structure of a spatial neural network of same size and density? In other words, we want to verify whether functional networks are rescaled versions of the spatial networks that generated the observed dynamics. In our framework this analysis is possible since we know how to create spatial networks of virtually any size and density (Section~\ref{sec:neuralnet}). Consequently, we generated new sets of 20 spatial networks of sizes 40, 50, \ldots, 100 and same densities as before (lower, intermediate, and higher) to properly perform comparisons with the respective functional networks. Simulation results are shown in the next section.

\section{Results and Discussion}
\label{sec:results}

\begin{figure*}[!t]
	\centering
	\includegraphics[width=1.0\textwidth]{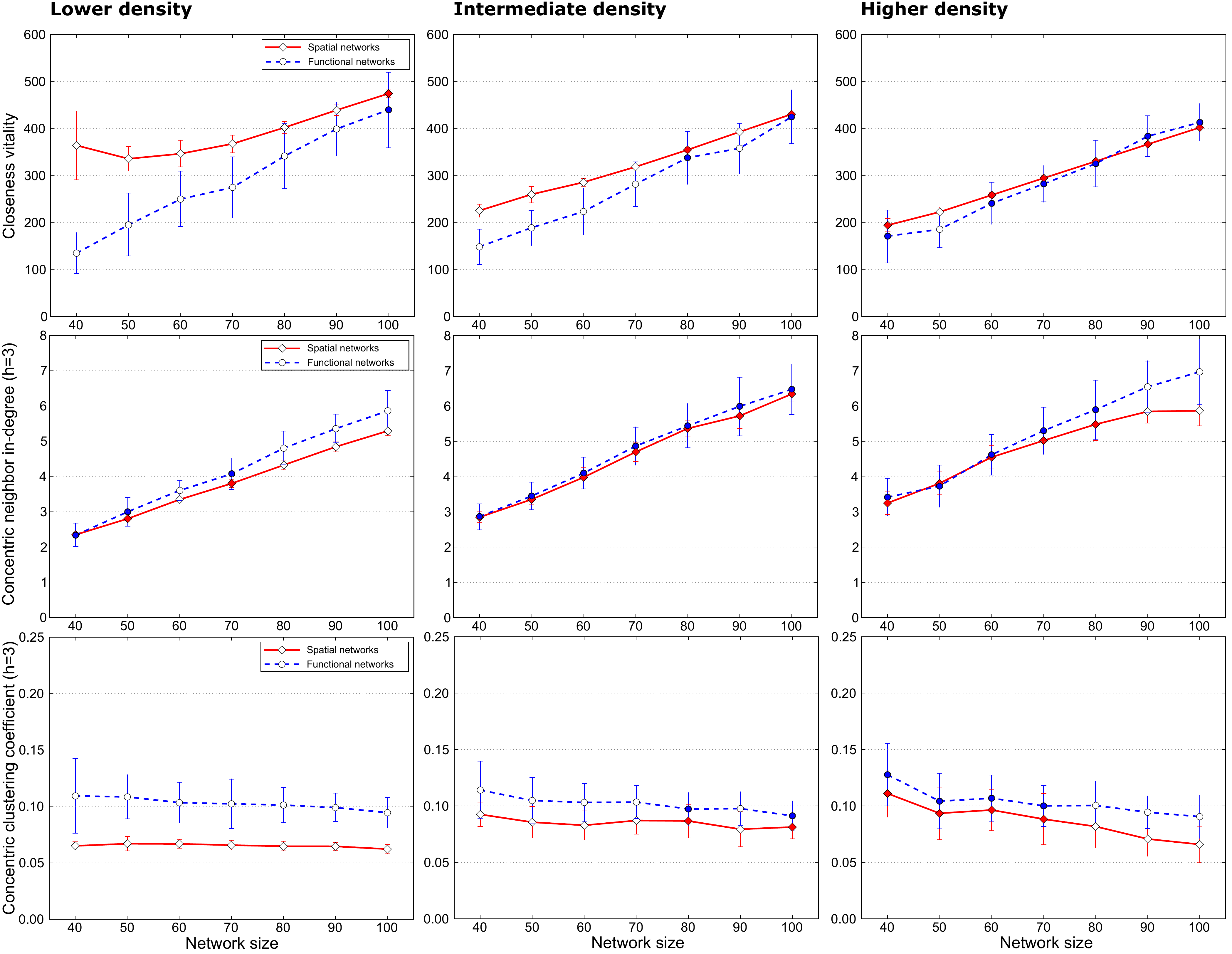}
	\caption{Comparison between functional and spatial networks with respect to three measurements: closeness vitality (first row), concentric neighbor in-degree at level $h=3$ (second row) and concentric clustering coefficient at $h=3$ (third row). Each column refers to one specific network edge density: lower (first column), intermediate (second column) and higher (third column). Each dot refers to an average computed over all network realizations (vertical bars indicate standard deviations). Filled dots mean that, for a given network size, it is not possible to reject the null hypothesis of equal averages (to reject the null hypothesis we considered the t-test with p-value < 0.01).}
	\label{fig:best}
\end{figure*}

Results are mainly discussed in terms of network density (lower, intermediate, and higher) and sample size (40 to 100). We also take into consideration the type of network measurements under analysis, which can be classified into \emph{local}, \emph{concentric} and \emph{global} measurements according to the type of information considered in calculations. Local measurements employ information from the immediate vicinity of nodes (e.g, degree, clustering coefficient and assortativity). Those classified as concentric consider the generalization of node neighborhoods to further distances. Global measurements need to take into account the entire network topology in order to compute minimum paths (e.g., efficiency, betweenness and vitality). We first introduce results associated with the measurements that best approximate functional and spatial networks and gradually show which measurements are not able to produce good estimations.

\begin{figure*}[!t]
	\centering
	\includegraphics[width=1.0\textwidth]{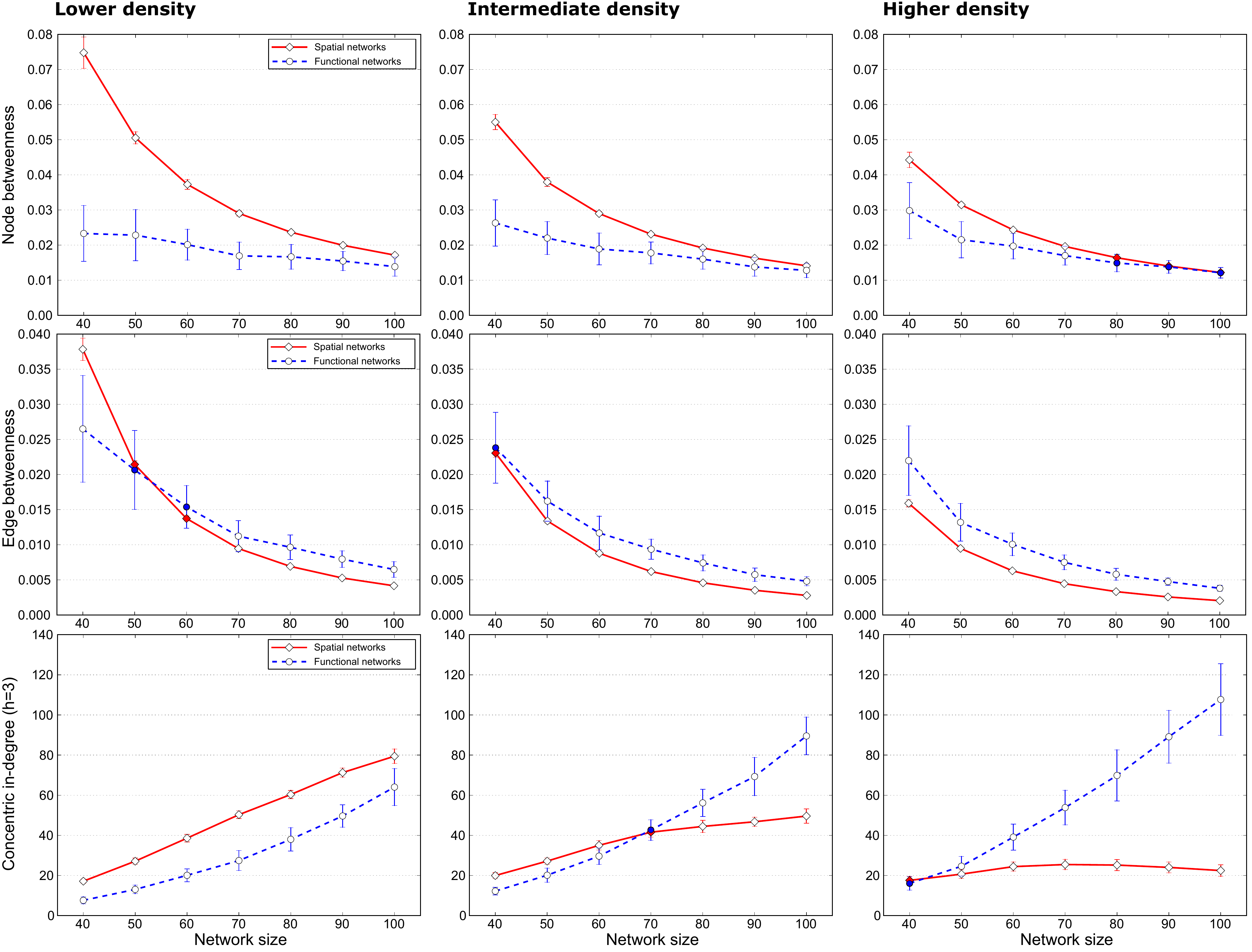}
	\caption{Comparison between functional and spatial networks, analogous to Fig.~\ref{fig:best}, now considering node and edge betweenness centrality, as well as the concentric in-degree at level $h=3$.}
	\label{fig:spec1}
\end{figure*}

\subsection{Best Approximations}
\label{sec:best}

Three measurements tend to be very similar when computed for functional and spatial networks. The leading one is the concentric neighbor in-degree at level $h=3$ (Fig.~\ref{fig:best}, second row). Approximations are very good for intermediate densities, with slight deviations when the largest network sizes are employed for other densities. Notice that the best approximations are indicated by filled dots in Fig.~\ref{fig:best} (details in the figure legend). Results were strikingly similar to the case of out-degrees (concentric neighbor out-degree at $h=3$, results not shown).

Closeness vitality also shows very good approximations (Fig.~\ref{fig:best}, first row), mainly for the case of higher densities. As densities decrease approximations get worse, although the largest network sizes are still able to accurately estimate the vitality of spatial networks in functional samples. Fig.~\ref{fig:best} also includes results for the concentric clustering coefficient at $h=3$ (third row). Again, estimations are better when higher densities are adopted. Notice that no local measurement is among the best approximations (see below), only global and concentric. Nevertheless, results are sensitive to network density.

\subsection{More Specific Approximations}

Other measurements are even more sensitive to sample size and network density, although still showing in specific conditions relevant approximations. It is the case, for instance, of edge/node betweenness centralities (Fig.~\ref{fig:spec1}). Although both measurements are intimately related, they show opposite behavior: node betweenness is best approximated for higher densities and larger network sizes, whereas edge betweenness shows good estimations when lower/intermediate densities and smaller network sizes are adopted.

A distinctive behavior is shown by the concentric in-degree at level $h=3$ (Fig.~\ref{fig:spec1}). Average values for functional networks grow steadily, whereas for spatial networks they tend to decay after a peak. Nevertheless, good estimations are shown in specific points (sizes 70 and 40 for intermediate and higher densities, respectively). Very similar results were observed for the out-degree analogous (concentric out-degree at $h=3$, results not shown). Qualitatively similar results were observed for the concentric number of nodes at $h=3$, where specific points of approximation were observed for lower and intermediate densities (sizes 90 and 60, respectively -- results not shown).

Other measurements highly diverge as network sizes increase, with values for functional networks steadily growing whereas for spatial networks they tend to decrease. These measurements are: concentric in-/out-degrees, concentric neighbor in-/out-degrees, and concentric number of nodes, all at $h=4$. Nevertheless, it is still possible to get good approximations with size 40 and lower densities (results not shown). Level $h=4$ is therefore a mesoscopic turning point between functional and spatial networks: the former tend to have significantly higher populations (nodes and edges) at this level. In other words, concentric analyses are more limited (in the sense of increasing $h$) in spatial networks than in functional ones.

\begin{figure*}[!t]
	\centering
	\includegraphics[width=1.0\textwidth]{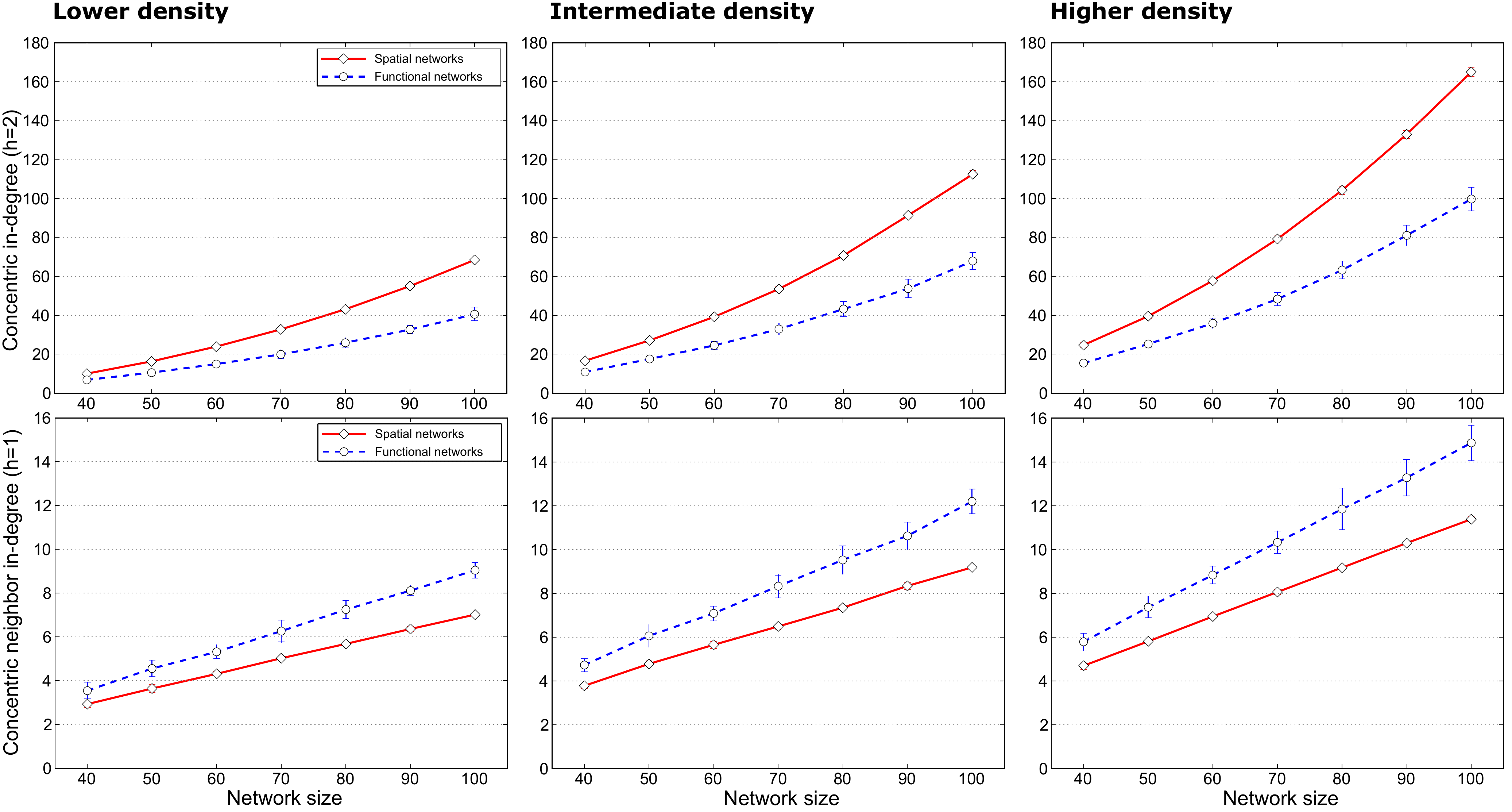}
	\caption{Analogous to Figs.~\ref{fig:best} and~\ref{fig:spec1}, now considering the concentric in-degree at level $h=2$ and the concentric neighbor in-degree at $h=1$.}
	\label{fig:spec2}
\end{figure*}

Another type of result also concerns concentric measurements. In this case, average values also tend to diverge for increasing sizes and densities. Nevertheless, for both network types there is a tendency of increasing values (Fig.~\ref{fig:spec2}, see also the very low standard deviations). In some cases, values for spatial networks are steadily higher than those for functional samples (concentric in-degree at $h=2$ -- first row of Fig.~\ref{fig:spec2} -- very similar to the concentric out-degree at the same level -- not shown). Nevertheless, good approximations can be seen when the lowest network size is applied (but averages are still statistically different). In other cases functional values are constantly higher than spatial ones, as for the concentric neighbor in- and out-degrees at levels 1 and 2 and maximum in- and out-degrees (bottom row of Fig.~\ref{fig:spec2}, only one of those measurements is shown). The best approximations are again obtained for the smallest network size.

Global and concentric measurements are again able to significantly estimate values observed in spatial networks, although in more restricted conditions than those highlighted in Section~\ref{sec:best}. Moreover, notice that variations between in- and out-degrees show extremely similar results for all measurement types. One possible cause is that directed edges have symmetric probabilities of creation in the spatial network model. Therefore opposing edge directions tend to be created in a (global) balanced way. Also notice that variations between in- and out-degrees must indeed yield the same results for functional networks since they have undirected edges.

\subsection{Severely Limited Approximations}

\begin{figure*}[!t]
	\centering
	\includegraphics[width=1.0\textwidth]{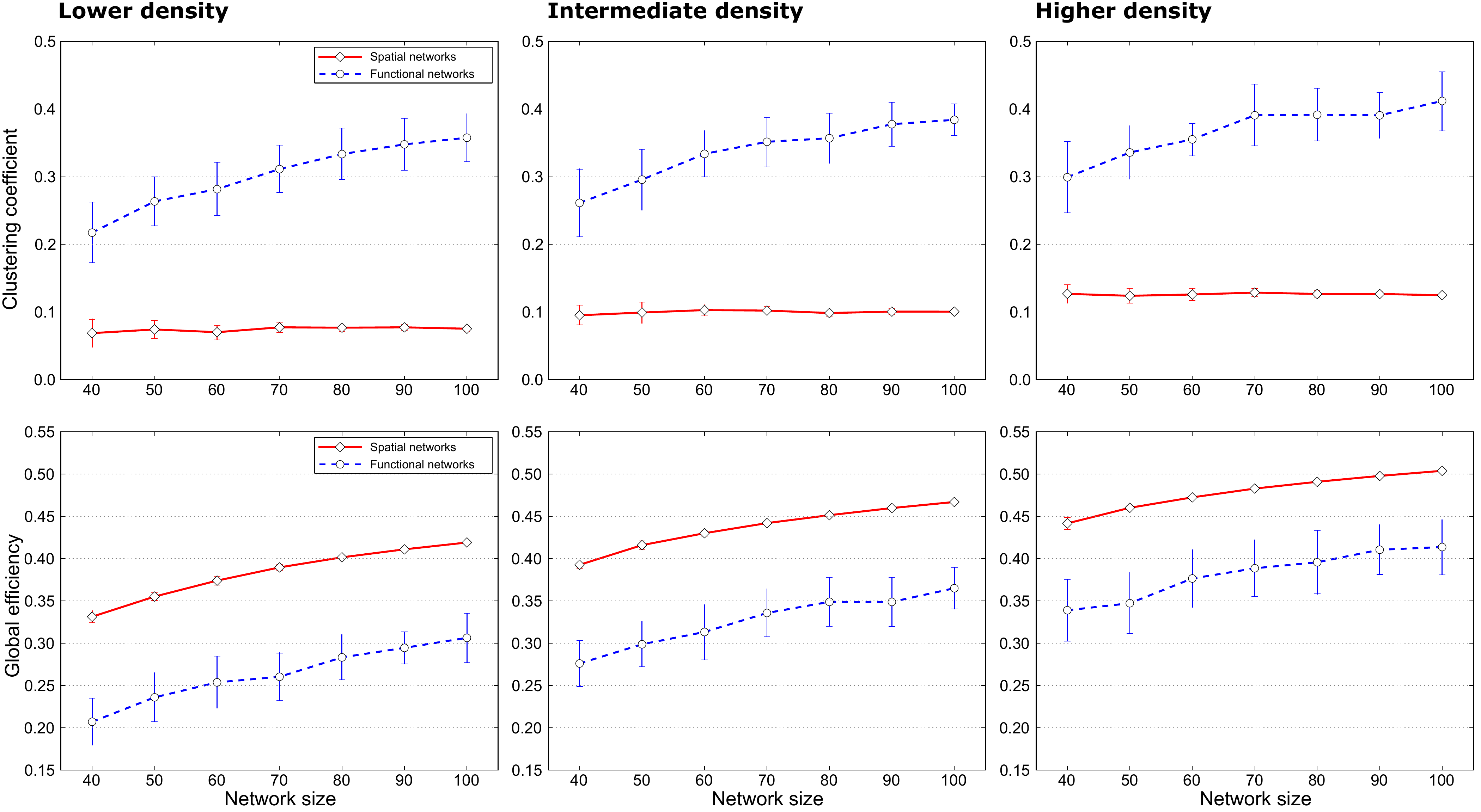}
	\caption{Analogous to Figs.~\ref{fig:best}--\ref{fig:spec2}, now considering the clustering coefficient and global efficiency.}
	\label{fig:limited}
\end{figure*}

The remaining measurements tend to be very different among spatial and functional networks. For instance, the clustering coefficient of functional samples is much higher that those for spatial networks (Fig.~\ref{fig:limited}, first row). Similar results were observed for the concentric clustering coefficient at $h=2$ and 4 (not shown). The relative difference in global efficiency between network types tends to be preserved as sizes and densities increase, although approximations are low quality in general (Fig.~\ref{fig:limited}, second row). Lengths of shortest paths are much smaller in spatial networks, with even worse approximations for lower densities (not shown). The reason for this behavior is that it is not uncommon to see isolated nodes in functional networks (this is also one of the possible causes for higher efficiencies appearing in spatial networks). The concentric number of nodes at $h=2$ (not shown) steadily grows like measurements in Fig.~\ref{fig:spec2}, although values between network types are very different even for the best cases (small network sizes). Finally, all assortativity types are poorly represented in both networks, with highly varying averages (i.e., no tendency observed) and very large standard deviations (results not shown). Values for functional networks tend to be slightly higher than those for spatial networks (in general $0.1$ against $0.0$). It is reasonable to say that, although estimations are not sufficiently precise, degree correlations (assortativity mixing) in both network types tend to be very low or even absent.

\section{Concluding Remarks}
\label{sec:concl}

Our modeling approach shows that it is possible to precisely estimate network measurements of spatial networks by just observing functional samples inspired by EEG experiments, although specific conditions of sampling size and density must be met. Local network measurements tend to be poorly estimated, whereas good approximations were observed for several global and concentric measurements. We thoroughly modeled and planned simulations in order to be able to perform a wide evaluation of functional samples in several different circumstances.

Beyond those contributions, another important goal is to adapt established computer science tools (in this case, spiking neural networks) on contemporary network problems. Although computer science has been playing a crucial role in the development of relevant network-based methodologies for diverse tasks (e.g., image segmentation~\cite{Felzenszwalb2004}), they are only sparsely integrated with the modern approach emerging from the highly multidisciplinary complex network field~\cite{Newman2010}.

Further developments of this work shall incorporate more realistic simulations (e.g., synaptic plasticity) and other types of analyses considering, for instance, nonlinear correlations between time series and network resilience under perturbations. All in all, besides specific numerical results, we report in this paper a general and robust framework for simulating and analyzing the confidence of functional samples that can be further adapted to other neuroimaging techniques (e.g., magnetoencephalography).

\bibliographystyle{IEEEtran}
\bibliography{lucas}

\end{document}